\documentclass{IEEEtran}
\usepackage{cite}
\usepackage{amsmath,amssymb,amsfonts}
\usepackage{graphicx}
\usepackage{textcomp,nicefrac}
\usepackage[caption=false]{subfig}

\usepackage{float}
\usepackage{subfig} 
\def\BibTeX{{\rm B\kern-.05em{\sc i\kern-.025em b}\kern-.08em
T\kern-.1667em\lower.7ex\hbox{E}\kern-.125emX}}
\begin{document}
\title{Invertible Low-Dimensional Modelling of X-ray Absorption Spectra for Potential Applications in Spectral X-ray Imaging}
\author{Raziye Kubra Kumrular and Thomas Blumensath 
\thanks{

R. K. Kumrular and T. Blumensath are with the ISVR Signal Processing and
Audio Hearing Group, University of Southampton, Southampton SO17 1BJ, U.K.
(e-mail: r.k.kumrular@soton.ac.uk )
}
\thanks{ }}

\maketitle

\begin{abstract}

X-ray interaction with matter is an energy-dependent process that is contingent on the atomic structure of the constituent material elements. The most advanced models to capture this relationship currently rely on Monte Carlo (MC) simulations. Whilst these very accurate models, in many problems in spectral X-ray imaging, such as data compression, noise removal, spectral estimation, and the quantitative measurement of material compositions, these models are of limited use, as these applications typically require the efficient inversion of the model, that is, they require the estimation of the best model parameters for a given spectral measurement. Current models that can be easily inverted however typically only work when modelling spectra in regions away from their K-edges, so they have limited utility when modelling a wider range of materials. In this paper, we thus propose a novel, non-linear model that combines a deep neural network autoencoder with an optimal linear model based on the Singular Value Decomposition (SVD). We compare our new method to other alternative linear and non-linear approaches, a sparse model and an alternative deep learning model. We demonstrate the advantages of our method over traditional models, especially when modelling X-ray absorption spectra that contain K-edges in the energy range of interest.

\end{abstract}


\section{Introduction}
\label{sec:introduction}

\IEEEPARstart{X}{-ray} Computed Tomography (XCT), which generates volumetric images based on measurements of X-ray transmission through an object, is a versatile imaging technique with applications in industry, security, medicine, and scientific investigation \cite{blumensath2015non,mery2015computer}. X-ray transmission is a function of X-ray energy, and the measurement of this dependency can be of significant importance in many applications. We are here interested in building models of this dependency that can help achieve this by allowing us to remove measurement noise, compress measurement data and constrain the estimation from limited measurements. In these applications, using models that both constrain the measurement whilst also allowing for easy estimation of the model parameters is crucial. 

 Whilst the physical interaction between photons and material can be modelled explicitly via very accurate Monte Carlo (MC) simulations\cite{busi2019enhanced, nazemi2021monte}, these models do not allow for simple model inversion. We thus here develop models with few parameters (so-called low-dimensional models) that are easy to invert, that is, that allow us to easily compute optimal parameters for a given X-ray spectral observation. These models can then be used to create a parameterised function as a computational tool for spectral data analysis that has a range of significant applications in X-ray imaging. For example, traditional XCT reconstruction algorithms that ignore energy dependence produce image artefacts which can be removed when using invertible low-dimensional models \cite{blumensath2015non}. Furthermore, low-dimensional models are crucial to remove measurement noise or constrain the ill-conditioned inverse problems that arise in several spectral imaging methods \cite{kumrular2022multi, busi2019enhanced}. 
 
Our work here is particularly motivated by our interest in measuring the spatial distribution of X-ray absorption spectra using commonly available lab-based X-ray tomography systems. There are several approaches to this. X-ray sources found in these systems generate X-ray photons with a range of energies (the X-ray source spectrum $I_{0}(E)$), though the X-ray detector does not normally differentiate different energy levels. To estimate absorption spectra, Dual-Energy CT uses two source spectra to allow spectral estimation \cite{alvarez1976energy} by utilising a two-parameter linear absorption spectral model. In Multi-Energy computed tomography (MECT), also called spectral X-ray tomography, spectrally resolved measurements are taken using photon counting detectors (PCD), though this comes at the cost of additional hardware requirements, a significant decrease in measurement speed, a significant increase in measurement noise as well as an increase in computational loads associated with the increase in measured data \cite{nik2013optimising}. In all of these applications, a more accurate invertible low-dimensional model of the X-ray absorption spectra is of significant interest, especially when imaging a wide range of materials. 

The attenuation of an X-ray beam with photons of a single energy ($E$) travelling along a path through an object is often modelled using the Beer-Lambert law:
\begin{equation}
I(E)=I_{0}(E)e^{-\int{\mu(x,E)} \ dx}\label{eq2}
\end{equation}
where $I(E)$ is the X-ray intensity measured by the detector, and $ I_{0}(E) $ is the X-ray intensity that would be measured by the detector without an object present. $ \mu(x,E) $ is the energy-dependent X-ray linear attenuation coefficient (LAC) at position $x$ along the X-ray beam and the integration is along the line of the  X-ray path through the object.

For X-ray energies below about 1.02 MeV, X-ray material interactions are due to three primary phenomena: Rayleigh scattering ($\mu_{R}(E)$); Compton scattering ($\mu_{C}(E)$); and the Photoelectric effect ($\mu_{P}(E) $) \cite{hendee2003medical}.  The total linear attenuation coefficient $ \mu(E)$ can thus be written as:
\begin{equation}
\mu(E)=\mu_{R}(E)+ \mu_{C}(E) +\mu_{P}(E) \label{eq3}
\end{equation}

\begin{figure} [h]
\center
\subfloat[]{{\includegraphics[width=0.8\linewidth]{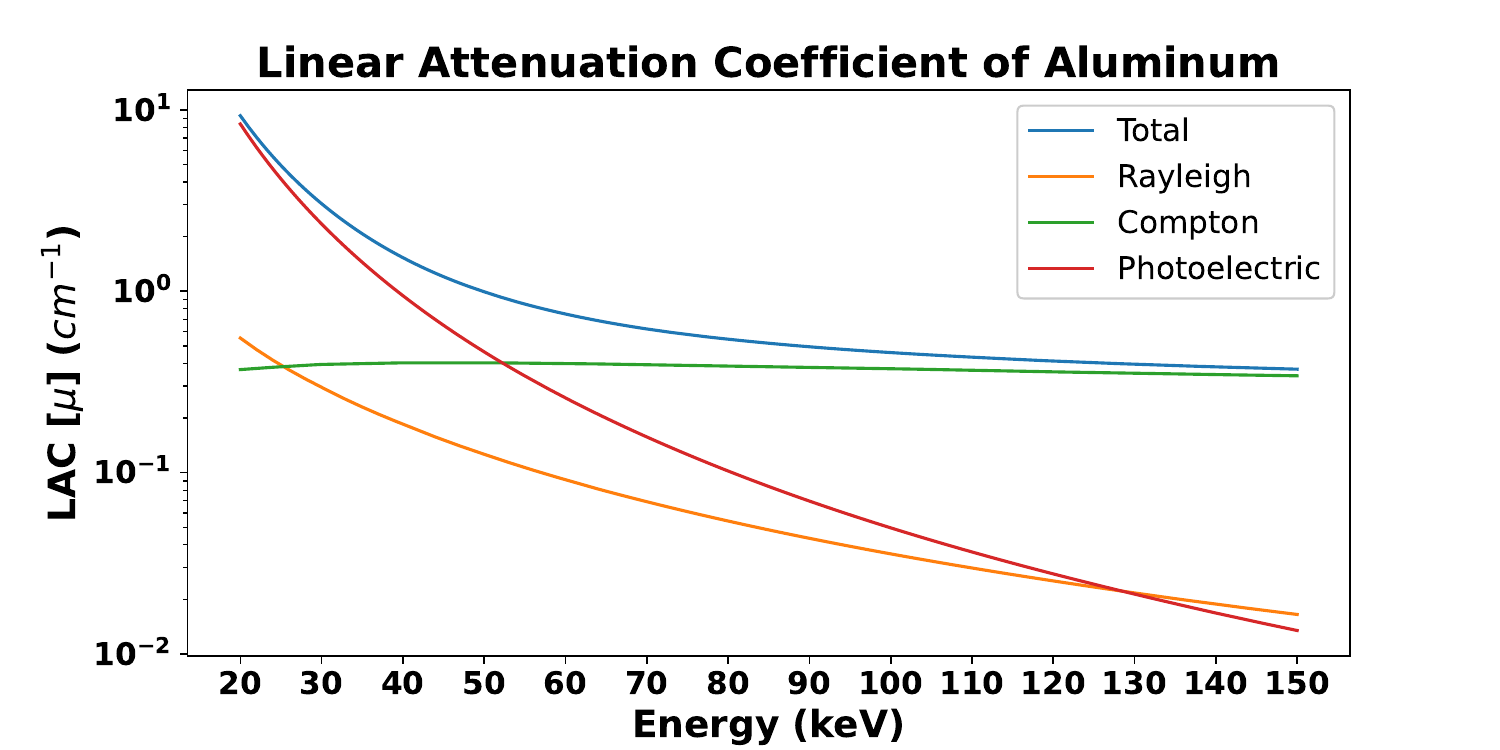}}  }
        \qquad
      \hfil
\subfloat[]{\includegraphics[width=0.8\linewidth]{images_pdf/Aluminum.pdf} }
\caption{Contributions of 3 types of phenomena in the selected energy range to the total linear attenuation coefficient of the materials: Rayleigh scattering, Compton scattering, and the Photoelectric effect. a) LAC of Aluminum b) LAC of Iodine. 
Data here is from the XCOM Photon Cross Sections Database \cite{berger2013xcom}, which was derived from fundamental quantum-physical principles.}%
\label{fig:cros_section}%
\end{figure}

Figure \ref{fig:cros_section} shows the total linear attenuation of Aluminum $(atomic-number(Z)=13)$ and Iodine $(Z=53)$ and the contribution of each of these interactions.
We here show the LAC as a function of energy, focusing on the energy range between 20 keV and 150 keV commonly used in lab-based X-ray systems. Of particular interest for our paper will be the step in the LAC due to the Photoelectric effect (as seen in Fig \ref{fig:cros_section}b for Iodine at 33.17 keV), which appears at the K-shell binding energy of the atom. This step is known as the K-edge of the element and is unique for each element \cite{hendee2003medical}.

The intrinsic dimensionality of the LACs using a linear principal component model has been studied previously in different settings \cite{bornefalk2012xcom, xie2020principal, tang2021conditioning}. We here instead investigate non-linear low-dimensional models of the X-ray absorption spectrum that work for all elements $(Z\leq 92)$ and over energy ranges found in typical lab-based tomography systems $(20keV\leq E\leq150keV)$.

\section{State of the art absorption spectrum modelling}
\label{sec:State of the art absorption spectrum modelling}

\subsection{Representation of Linear Attenuation Coefficient with Linear Models}

\label{subsec:Representation of Linear Attenuation Coefficient}

Different X-ray absorption models have been proposed in the literature. These models typically  assume absorption spectra to be a linear combination of two or more basis functions, which are assumed to be independent of the material \cite{alvarez1976energy, lehmann1981generalized}.

\subsubsection{Photoelectric-Compton Basis (PCB) model}
The first model is based on \eqref{eq3}. Rayleigh scattering is negligible. \cite{martz2016x}. The Photoelectric-Compton Basis (PCB) model thus only uses basis functions to represent the Photoelectric effect and Compton scattering which are assumed to be material invariant.
This holds for energies far from the K-edge energy of a material (see Fig. \ref{fig:cros_section}a), where the Photoelectric absorption and Compton scattering phenomenon can be approximated as \cite{nik2013optimising,devadithya2021enhanced}: 

\begin{equation}
\mu(E)=a_{p}f_{p}(E) + a_{c}f_{KN}(E) 
\label{PCB_model}
\end{equation}

where  $ f_{p}(e)$ and $f_{KN}(e) $ are functions of energy only and capture the energy dependence of the Photoelectric absorption and Compton scattering. $a_{p}$ and $a_{c}$ on the other hand are parameters that are independent of energy and instead only vary with the material (they are functions of the electron density $(\rho_{e})$ and the atomic number of the material).  $a_{p}$  and $a_{c}$ are thus two parameters that can be used to fit this linear two-dimensional model to data \cite{alvarez1976energy, azevedo2016}. For a single material, they can be derived as functions of $\rho_{e}$ and $Z$ as:

\begin{equation}
\label{aP}
\begin{split}
& a_{p}=\rho_{e}C_{P}{Z^{m}}\\  
& a_{c}=\rho_{e},
\end{split}
\end{equation}
where $ C_{P} $ =  $9.8$x$10^{-24}$ \cite{mccullough1975photon} is a constant, and m= 3.8 was determined experimentally.

The energy dependence of the Photoelectric effect is approximated by $f_{p}(E)={{1}/{E^{n}}}$ and it is possible to approximate the energy dependence of Compton scattering using the Klein-Nishina function (\ref{KN})\cite{macovski1983medical}.

\begin{equation}
\label{KN}
\begin{split}
f_{KN}(E) =&\frac{1+\alpha}{\alpha^{2}} \left[ \frac{2(1+\alpha)}{(1+ 2\alpha)}- \frac{1}{\alpha}\ln(1+ 2\alpha) \right] \\
& +\frac{1}{2\alpha}\ln(1+ 2\alpha) -\frac{1+3\alpha}{(1+ 2\alpha)^{2} } 
\end{split}
\end{equation}

where $\alpha $ is ${E}/{E_{e}} $ and  $E_{e} $  $\approx$ 511 keV denotes the rest mass energy of an electron. This two-dimensional linear model is suitable for low atomic number materials $(Z<18)$ that do not have a K absorption edge in the range of energies considered, though the approximation error increases close to the K-edge as well as for higher energies \cite{alvarez1976energy,champley2019method}.

\subsubsection{Material-basis (MB) model}

The second linear model uses material-basis (MB) functions, which are two or more LAC functions taken from previously chosen reference materials\cite{lehmann1981generalized}. This model is particularly popular in medical imaging, where the imaged object can be modelled using a basis function for the LAC of bone and one for soft tissue (or water)\cite{devadithya2021enhanced}. However, it is hard to express a wider range of materials with just two reference materials in the MB model, and this model does not provide a direct estimate of the electron density and effective atomic number \cite{devadithya2021enhanced, eger2011learning, busi2019enhanced} 

\subsubsection{Learned linear representations}
\label{subsec:Learned basis representation}
Basis functions for low-dimensional modelling of X-ray absorption spectra can also be learned from training data. This can be done using the singular value decomposition (SVD), which also gives an estimate of the approximation error that can be achieved. The SVD computes the best linear approximation to a given training dataset in the mean squared error sense for a given size of subspace. There is a close relationship between SVD and principal component analysis (PCA) \cite{wall2003singular}, which has been used several times to derive low-dimensional linear models \cite{bornefalk2012xcom, xie2020principal, tang2021conditioning}.

For materials without K-edge in the energy range, it has been found that SVD models provide good approximations to LACs using two basis functions \cite{lehmann1986energy, busi2019enhanced,champley2019method}. Furthermore, the learned basis functions are very similar to $\mu_{p}(E)$ and $\mu_{c}(E)$ \cite{busi2019enhanced,champley2019method}. 
However, these models no longer work close to a K-edge \cite{busi2019enhanced, lehmann1986energy}, though increasing the number of basis functions naturally has been found to increase performance also in these cases \cite{eger2011learning}.

\subsection{Non-linear models}
\label{subsec:Non-linear models}
Given the inability of low-dimensional linear models to capture K-edges in absorption spectra, non-linear models might be suitable alternatives. As there are no known analytic models that capture the change in absorption around the K-edge of all materials in a succinct parameterization, learned non-linear models are a viable alternative. 

\subsubsection{Sparse Model}
We have already introduced the idea of using a material basis function model. It is possible to include a basis function for each material in the periodic table, but as a linear model, this would require us to fit many coefficients. Instead, to derive models with few non-zero coefficients when using a larger set of material basis functions, sparse models can be used \cite{yaghoobi2016fast}. Whilst the generative model here is still linear (i.e. a spectrum is modelled as a linear computation of basis function), the estimation of the weights now becomes a non-linear process. To get a low-dimensional model, the basic assumption then is that a given spectrum represents a material that is a combination of a few elements.  
Sparse models have been suggested as a complement to traditional regression methods for better identification of spectra in Raman spectroscopy \cite{wu2014sparse}, though to the best of our knowledge, have not yet been used to model X-ray absorption spectra.

\subsubsection{Neural Network based Models}
Recent advances in deep learning now allow the estimation of complex non-linear relationships in complex data.
A suitable model for our purpose is an autoencoder, which is a deep neural network that can learn a non-linear low-dimensional representation \cite{routh2021latent}.  
The network consists of two main components; a non-linear encoder, which compresses the input into a latent space representation, and a non-linear decoder which reconstructs the data from the low-dimensional representation \cite{goodfellow2016deep}. For a single material, autoencoders have already demonstrated the ability to capture fine detail in the absorption spectrum around K-edge energies \cite{routh2021latent}. 
To increase robustness and to incorporate noise suppression, an autoencoder is often trained as a denoising autoencoder (DAE), where the difference in training is that the input to the encoder is corrupted by noise during training.

\section{Material and Methods}
\label{sec:material and methods}

In this paper, we hypothesize that a single non-linear low-dimensional latent representation will allow us to model the X-ray absorption spectra of all elements, including those that have a K-edge in the energy range of interest. As low $Z$ materials without a K-edge in the energy range under investigation are already well approximated using two linear basis functions, we propose to model the difference between a given spectrum and an optimal two-dimensional linear approximation. 

\subsection{Proposed Non-linear Model}
\label{subsec: Proposed Approach}
\begin{figure}[h]
\centerline{\includegraphics[width=1\linewidth]{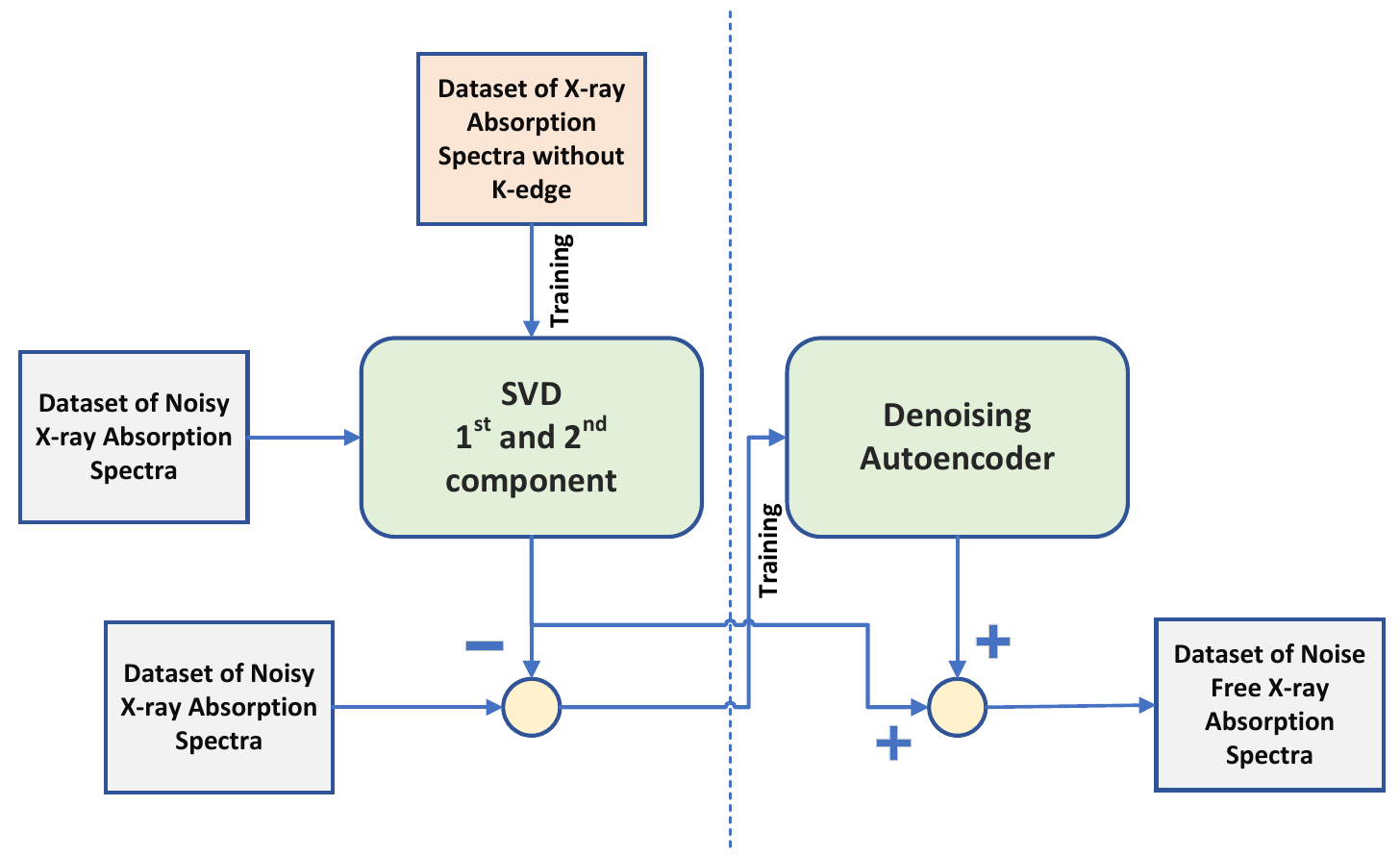}}
\caption{Schematic of the proposed approach. The SVD model is learned from the dataset of the X-ray absorption spectrum without the K-edge to have the $ 1^{st}$  and $ 2^{nd}$  components. The autoencoder is trained using the SVD model residual of the noisy X-ray absorption spectrum dataset with K-edges.
}
\label{fig: proposed approach}
\end{figure}

Our proposed model combines a non-linear autoencoder with a two-dimensional SVD-based representation as shown in Fig. \ref{fig: proposed approach}. 
The SVD learns the effect of the Photoelectric effect and Compton scattering for materials with low atomic numbers, where we do not have a K-edge in the energy range of interest. The autoencoder then uses a 3-node latent representation to try and model the deviation of the true spectrum from the linear model of materials that have a K-edge in the energy range.

\subsection{Low Dimensional Representation of  X-ray Absorption Spectrum with the Autoencoder}
\label{subsec:Low dimensional}

\begin{figure*}[h]
\centerline{\includegraphics[width=0.7\linewidth]{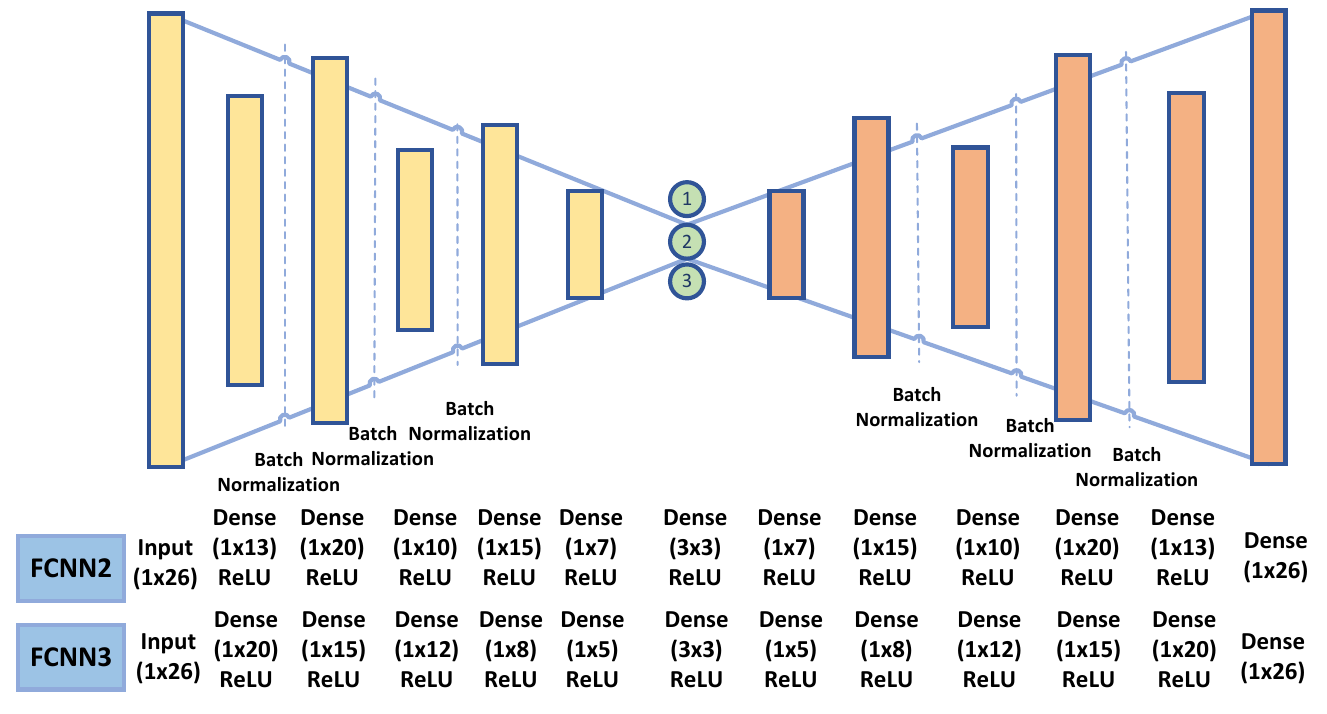}}
\caption{Illustration of created different deep learning models with the fully connected neural network. Selected model parameters of FCNN2 and FCNN3 are explained below in the illustration. }
\label{fig:FCNN}
\end{figure*}

\begin{figure*}[h]
\centerline{\includegraphics[width=0.7\linewidth]{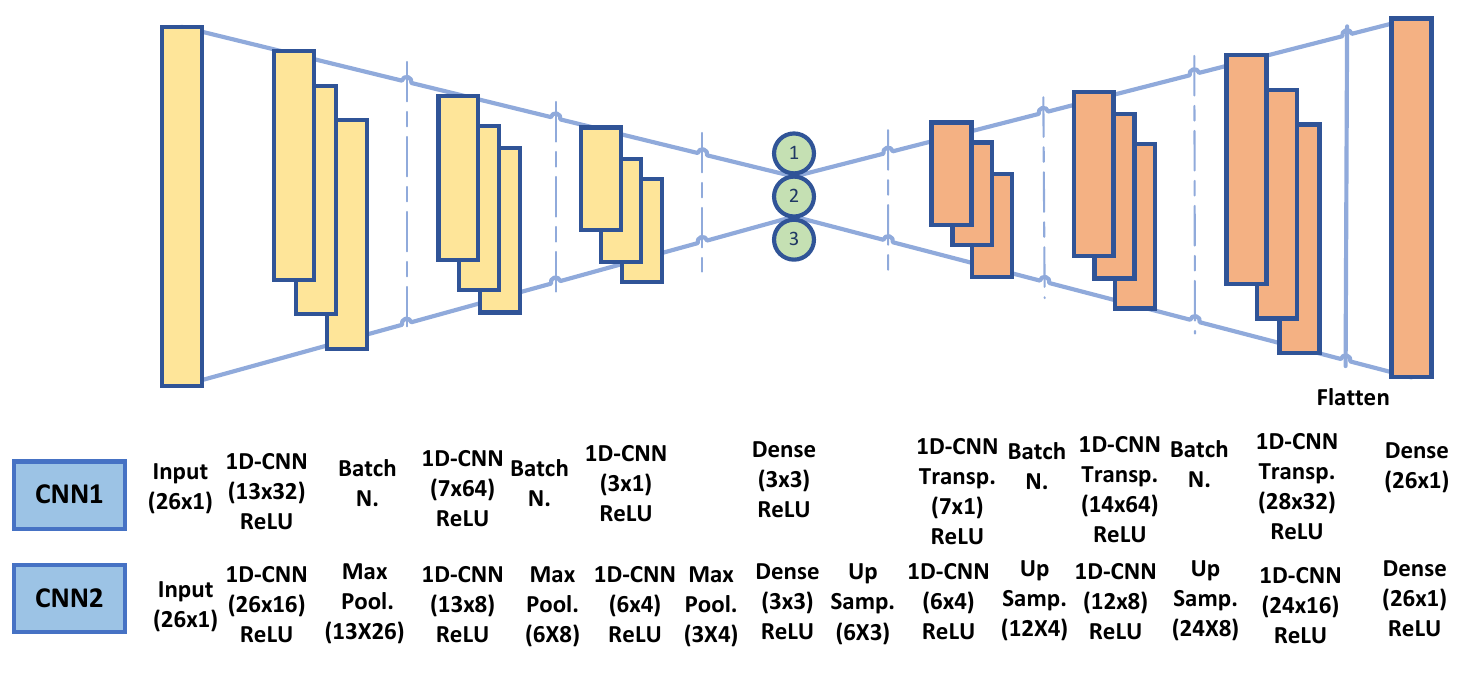}}
\caption{Illustration of created different deep learning models with the convolutional neural network. top: CNN1, bottom: CNN2 }
\label{fig:CNN}
\end{figure*}

There are different network architectures that can be used as the autoencoder in our model. We here compare two convolutional neural network (CNN) and three fully connected neural network (FCNN). The most basic FCNN simply consisted of a single input layer, the hidden (code) layer and an output layer with ReLU non-linearities in the input and output layers. The other four network architectures are shown in Fig.\ref{fig:FCNN} and Fig. \ref{fig:CNN}. These architectures all had 3 nodes in the latent space when used jointly with the two-component SVD model, or 5 nodes if used without an initial SVD
approximation.


The main difference between FCNN2 and FCNN3 is the layer structure. In the FCNN3, the number of nodes shrinks gradually in the encoder and expands gradually in the decoder, which is a regular layer structure for the autoencoder, whilst, for  FCNN2, the number of nodes in two consecutive layers first shrinks by about half before slightly expanding again in the next layer, a pattern that is repeated in the encoder and inverted in the decoder. Batch normalization is used to prevent overfitting.  The main difference between the two convolutional networks is that CNN1 uses strided convolutions, whilst in CNN2 we use max pooling. To apply our idea to datasets sampled at different energy levels (131 energy levels), the CNN2 architecture is modified by creating much deeper layers but using the same node number in the latent space.

\subsection{A Sparse Regularized Model for X-ray Absorption Spectrum}
As a comparison, we also implement a sparse model using a material basis function matrix. Let $Y$ be the X-ray absorption spectrum of a chemical mixture $(Y \epsilon$ $ R^{N})$, and $A$ $(A=a_1,a_2,...a_N )$ a matrix whose columns are the material basis functions of all elements of interest $(A \epsilon$  $R^{NxM} )$. To compute a sparse representation $X$, we solve the lasso problem:
\begin{equation} 
\underset{X}{\arg\min}
 \| \mathbf{Y} - AX \|_{2} +   \lambda \| X\|_{1}
\label{Fista}
\end{equation}
where we use the FISTA algorithm for optimisation. 
We here generate the material basis function matrix by using the LAC values for the 92 elements provided by the National Institute of Standards and Technology (NIST) database  \cite{NISTdatawebsite}. As the solution to the above lasso problem does potentially provide approximations of the data with more than 5 basis functions, for consistency with our 5 parameter model, we restrict the solution by selecting the 5 largest elements (in magnitude) of $X$ and then fitting these values by computing a least squares solution using only the selected 5 material basis functions.


\subsection{Traditional methods}
To compare the two non-linear models above to traditional linear and non-linear models, we furthermore implemented an SVD-based method, where we selected the largest 5 components to provide a 5-dimensional linear model. We also implemented our three autoencoder models without the initial 2-dimensional SVD model by extending the dimension of the hidden layer to 5. Thus, all our models could be used to fit 5 parameters into a spectrum.


\subsection{Dataset of the simulated X-ray absorption spectrum }
\label{subsec:dataset}
X-ray absorption spectra have been simulated using the linear attenuation coefficients of the 92 chemical elements with $Z\leq 92$. LAC values were obtained by multiplying MAC (Mass attenuation coefficient) with average mass densities obtained from the NIST \cite[Tables 1 and 3]{NISTdatawebsite}. 
The energy range of interest was chosen to be between 20 keV to 150 keV, which is the available source energy range found in many lab-based X-ray tubes. For computational efficiency, spectra were re-sampled into 26 equally sized energy bins, though similar results can be achieved with a finer energy resolution.

We generated a range of different datasets, consisting of combinations of between 1 and 5 different elements, with some datasets having pre-specified numbers of elemental spectra with K-edges. All datasets are summarised in Table \ref{table_materials}. Each mixture is generated by randomly choosing the elements (possibly with restrictions on the required numbers of K-edges) and then combining them by multiplying them by the standard elemental density for that material as well as a random scalar drawn from a uniform distribution in the range between 0 and 1. To consistent data, the datasets scaled with standardization after generating combined LACs. To train the de-noising autoencoders, Gaussian noise, with zero mean and 0.1 standard deviations, was added to generate a noisy version of each dataset.

We created various datasets to perform the proposed method and compare it with other methods. Table \ref{table_materials} shows the name of the generated datasets, where the subscript indicates the number of elements in each mixture in that dataset, e.g. each element in $D_{2E}$ consists of two randomly selected elements, as well as the number of the elements in each mixture that have K-edges, e.g. each element in $D_{2E,2K}$ contains two elements with K-edges in the energy range of interest (i.e. $(Z>42)$). The dataset containing 131 energy levels ($D_{2E,131}$) was generated the same way as the other datasets, the only difference being that it was quantized at every energy level.

\begin{figure} [h]
\center 
    \subfloat[]{{\includegraphics[width=0.8\linewidth]{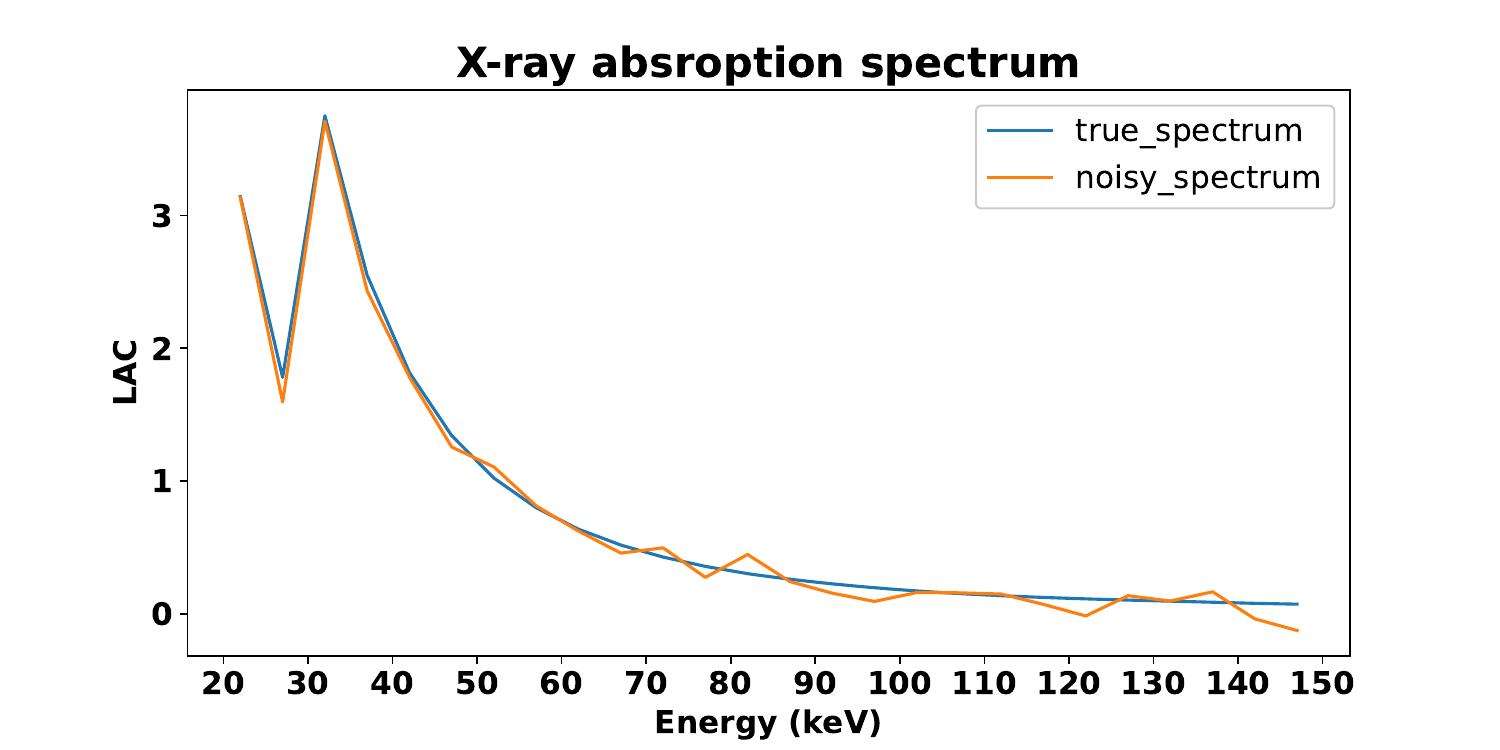} }} 
        \qquad
      \hfil
    \subfloat[]{{\includegraphics[width=0.8\linewidth]{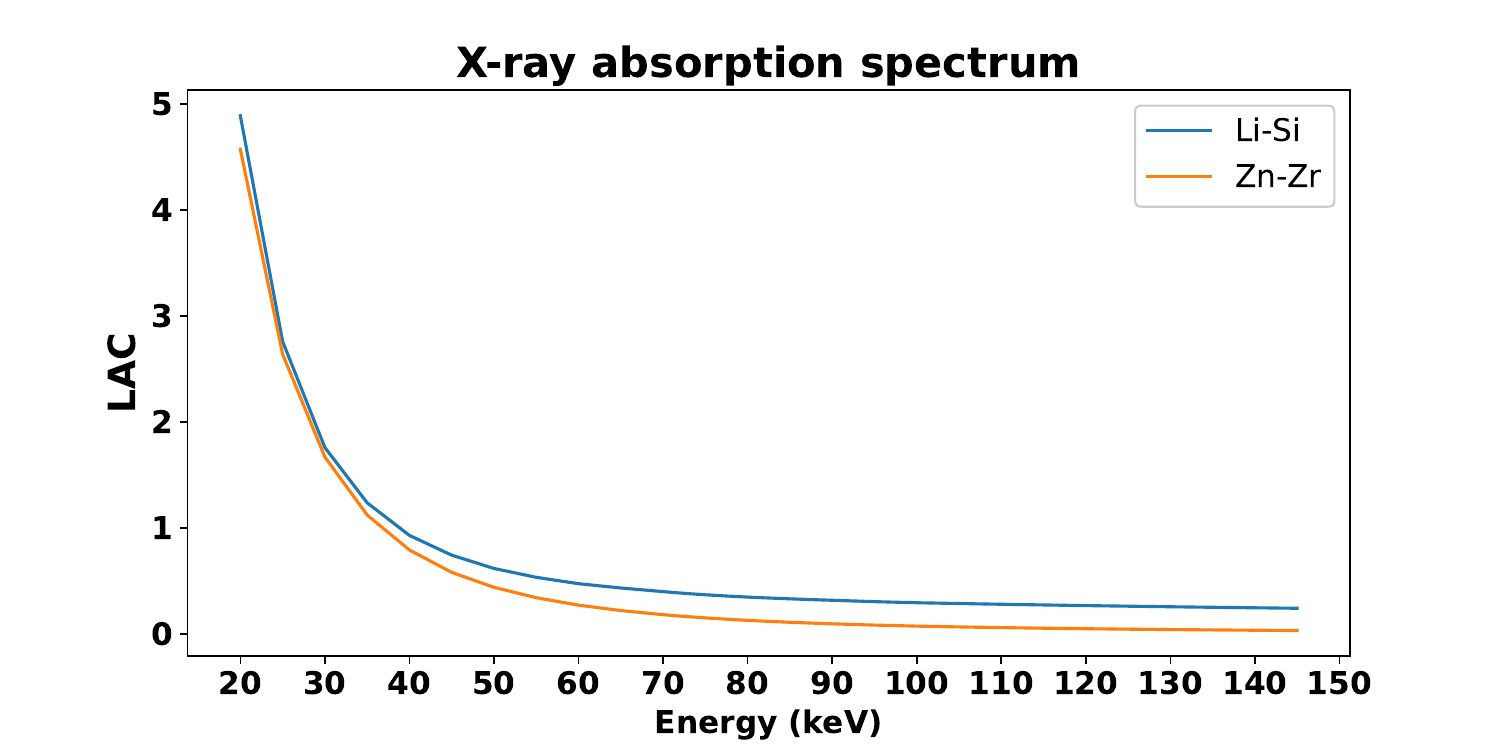}}}
\caption{Sample spectra from a) $(D_{2E})$ with and without noise and b) $D_{2E,0K}$ with elements that do not have any K-edge in the energy range (blue-line: Lithium and Silicon, orange-line: Zinc and Zirconium).}%
\label{fig:data_generation}%
\end{figure}

Example spectra are shown in Figure \ref{fig:data_generation}a  where we show noisy and noise-free spectra from $D_{2E}$, and Fig. \ref{fig:data_generation}b, where we show two example spectra from $D_{2E,0K}$.

\begin{table}
\centering
\caption{Material Compositions}
\label{table_materials}
\setlength{\tabcolsep}{3pt}
\begin{tabular}{|p{33pt}|p{37pt}|p{29pt}|p{93pt}|}
\hline
\textbf{Dataset Name}&
\textbf{Number of Elements}&
\textbf{Object Number  }&
\textbf{Description of Objects} 
 \\
\hline
$D_{2E}$& \centering2 & \centering 20000 & random K-edge in data \\
$D_{2E,131}$& \centering2 & \centering 20000 & random K-edge in sampled  data with 131 energy level \\
$D_{2E,0K}$ & \centering2 & \centering 1000 & without K-edge \\
$D_{5E}$ & \centering 5 & \centering 20000 & random K-edge in data\\
$D_{5E,0K}$ & \centering5 & \centering 1000 & without K-edge \\
$D_{2E,2K}$ &\centering2 &  \centering100   &  2 elements have K-edge\\ 
$D_{3E,0K}$ &\centering3 &  \centering100   &  without  K-edge\\ 
$D_{3E,1K}$ &\centering3 &    \centering100 & 1 element has K-edge  \\ 
$D_{3E,2K}$ &\centering3 &   \centering100  & 2 elements have K-edge \\ 
$D_{3E,3K}$ &\centering3 &  \centering100   & 3 elements have K-edge  \\ 
$D_{5E,5K}$ &\centering5 &  \centering100   & 5 elements have K-edge  \\ 
\hline
\multicolumn{4}{p{210pt}}{Here, the indices in parentheses represent the number of K edges in the X-ray absorption spectrums. The numbers in front of the letter E represent the number of elements in the objects. 131 indicates the number of samples in the energy for that dataset.}\\
\end{tabular}
\end{table}

\subsection{Loss function}
\label{subsec:Loss function}
To evaluate the performance of different methods, we use the normalised mean squared error (NMSE):

\begin{equation}
NMSE = \frac{{\| Y-\hat{Y}\|^2}}{{\|Y\|^2}},
\label{eq8}
\end{equation} 

where $Y $ is true X-ray absorption spectrum, and  $\hat{Y} $ is predicted X-ray absorption spectrum. $\| Y-\hat{Y}\| $ is the $l_{2}$ norm of the error between true spectrum and predicted spectrum, while $\| Y\|$ is the $l_{2}$  norm of the true spectrum.

\section{Experiments and Results}
\label{sec:results}

We test the sparse and machine learning-based non-linear models and compare them to the linear methods. In the rest of the paper, we referred to the proposed hybrid models as SVD/autoencoder and the autoencoder models with 5 nodes in the latent space layer as 5-dimensional autoencoders. We also fit an SVD model using the largest 5 components, which we call the 5-dimensional SVD. The sparse model, where we fit the largest 5 components after sparse decomposition is called the Fista model. 

All models that include one of the autoencoders were trained using the same parameters, using an Adam optimiser with a batch size of 64 and running for 300 epochs with a mean squared error loss function.
Unless otherwise stated, all autoencoder-based models were trained on the data-set $D_{2E}$, which was divided by random training (72\%), validation (20\%) and test (8\%). The validation set was used to validate the model performance during training.
For the SVD/autoencoder model, we trained the SVD and the autoencoder separately, starting by fitting the SVD using data without K-edges, namely $D_{2E,0K}$. We then trained the autoencoder in the  SVD/autoencoder model with the training data from $D_{2E}$, which also included simulated absorption spectra with K-edges. For the training of the autoencoder part of the SVD/autoencoder model, each spectrum was first projected onto the SVD subspace and the residual error was used to train the autoencoder. The output of the autoencoder was then added back to the approximation computed with the SVD model to provide the spectral approximation (as shown in Fig. \ref{fig: proposed approach}). 

We also used the same training dataset from $D_{2E}$ to fit the 5-dimensional SVD model. For the FISTA model, the sparsity parameter ($\lambda$) was optimised for optimal performance on the same dataset. As the SVD is known to provide the best linear low-dimensional approximation in the mean squared error sense, we do not report results for other linear models.

\begin{figure}[h]
\centerline{\includegraphics[width=1\linewidth]{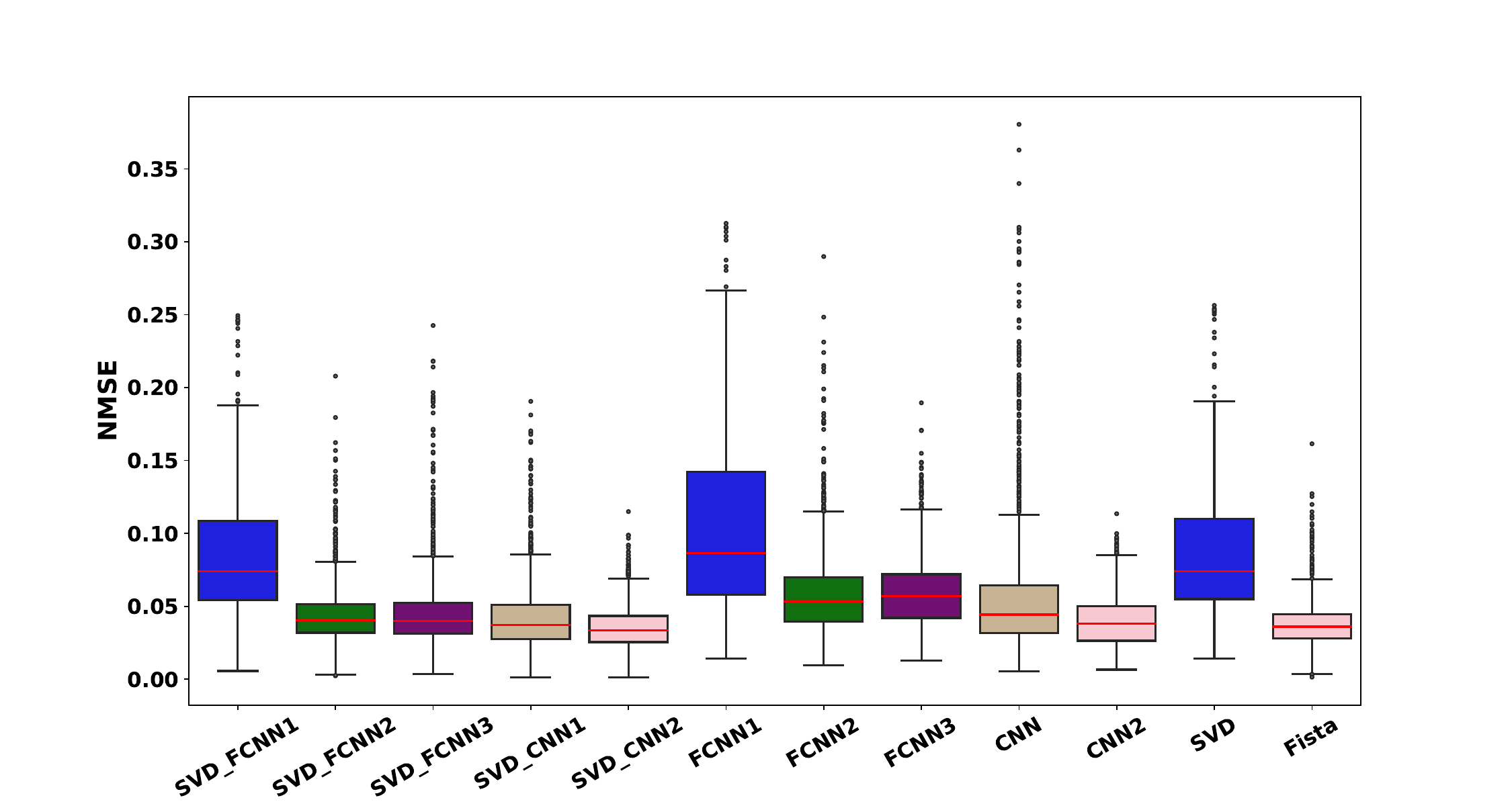}}
\caption{NMSE of the $D_{2E}$ test dataset. The two-material test dataset included 1600 X-ray absorption spectra. Each colour in the graph denotes SVD/autoencoder architectures and a corresponding 5-dimensional autoencoder network. The 5-dimensional SVD is shown in blue and Fista  in pink.
 }
\label{fig:two_element}
\end{figure}

After the training step, all models were tested on the test dataset of $D_{2E}$, and each model was evaluated with by plotting Box-whisker plots of the  MMSE for each spectrum in the test data. Fig. \ref{fig:two_element} shows results for the SVD/autoencoder models, the 5-dimensional autoencoder models, the 5-dimensional SVD model as well as the Fista model. From these results, we see that CNN2 performs better as the non-linear model, both on its own or in conjunction with the initial SVD projection. (Similar results were found when analysing other datasets (results not shown for brevity).) For the remainder of this paper, we thus only report the results for the CNN2-based models, the 5-dimensional SVD model and the Fista model.

\begin{figure}[h]
\centerline{\includegraphics[width=0.8\linewidth]{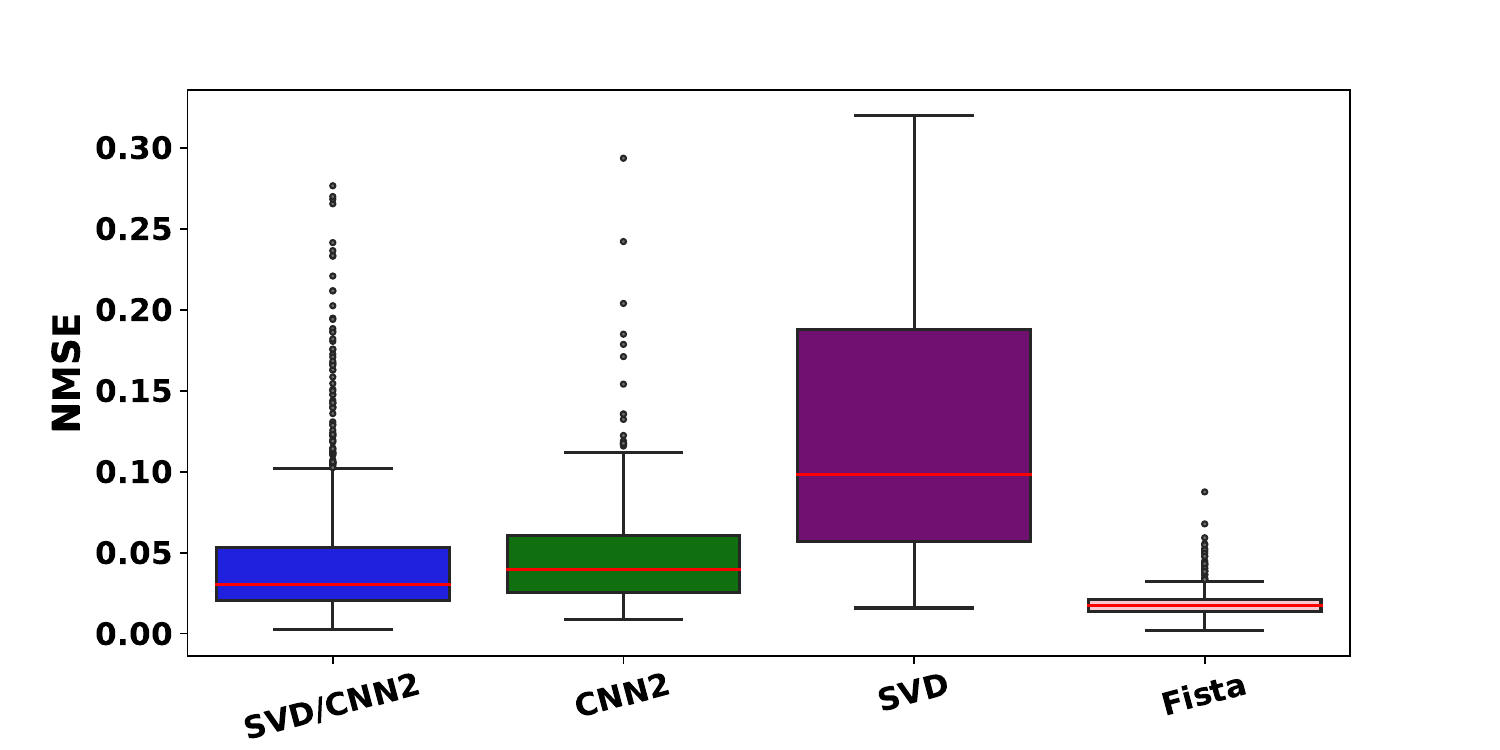}}
\caption{NMSE of the ($D_{2E,131}$) test dataset. This dataset included 1600 X-ray absorption spectra like $D_{2E}$ test dataset, and the only difference is the sampling number in the energy range. }
\label{fig:131 energy level}%
\end{figure}

To research the performance of our ideas on the dataset with finer energy resolution, we followed the same steps in the training and testing with other architectures. We focused on two different architectures that extended versions of CNN2 (have a better result than others) in this experiment. Figure \ref{fig:131 energy level} shows the result of the modified version of SVD/CNN2 and CNN2 along with Fista and 5-dimensional SVD results for $D_{2E,131}$ dataset. The average NMSE performance here is similar to that found for the $D_{2E}$ dataset.

\begin{figure}[h]
\centerline{\includegraphics[width=0.8\linewidth]{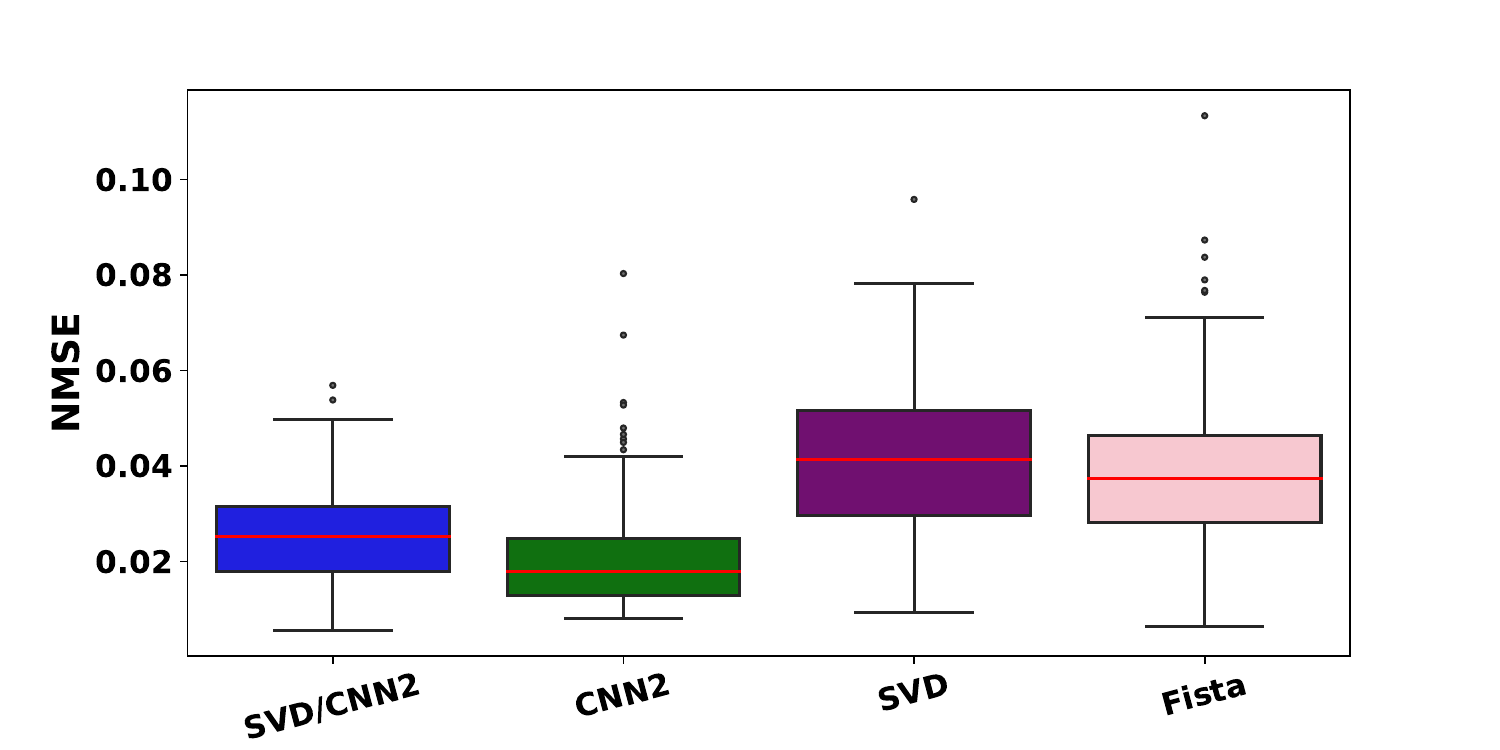}}

\caption{NMSE of the three-element material dataset ($D_{3E,0K}$) created without the K-edge.}
\label{fig:3element_without_K_edge}%
\end{figure}

To further demonstrate this, the modelling performance of our approach was also tested using the $D_{3E,0K}$ (see Fig. \ref{fig:3element_without_K_edge}) of 3 element mixtures without K-edge. For this dataset without K-edges, we again found that the SVD/autoencoder model no longer outperforms all other methods, and in fact, the 5-dimensional CNN2 now performed slightly better in terms of the mean of the NMSE errors. Crucially, the 5-dimensional SVD and Fista models showed almost the same performance. Of interest here is also the fact that the 5-dimensional SVD does not work as well as the non-linear models, which is likely due to the fact that the linear approximation used is not valid in energy ranges close to K-edges.

\begin{figure}
\center 
    \subfloat[]{{\includegraphics[width=0.8\linewidth]{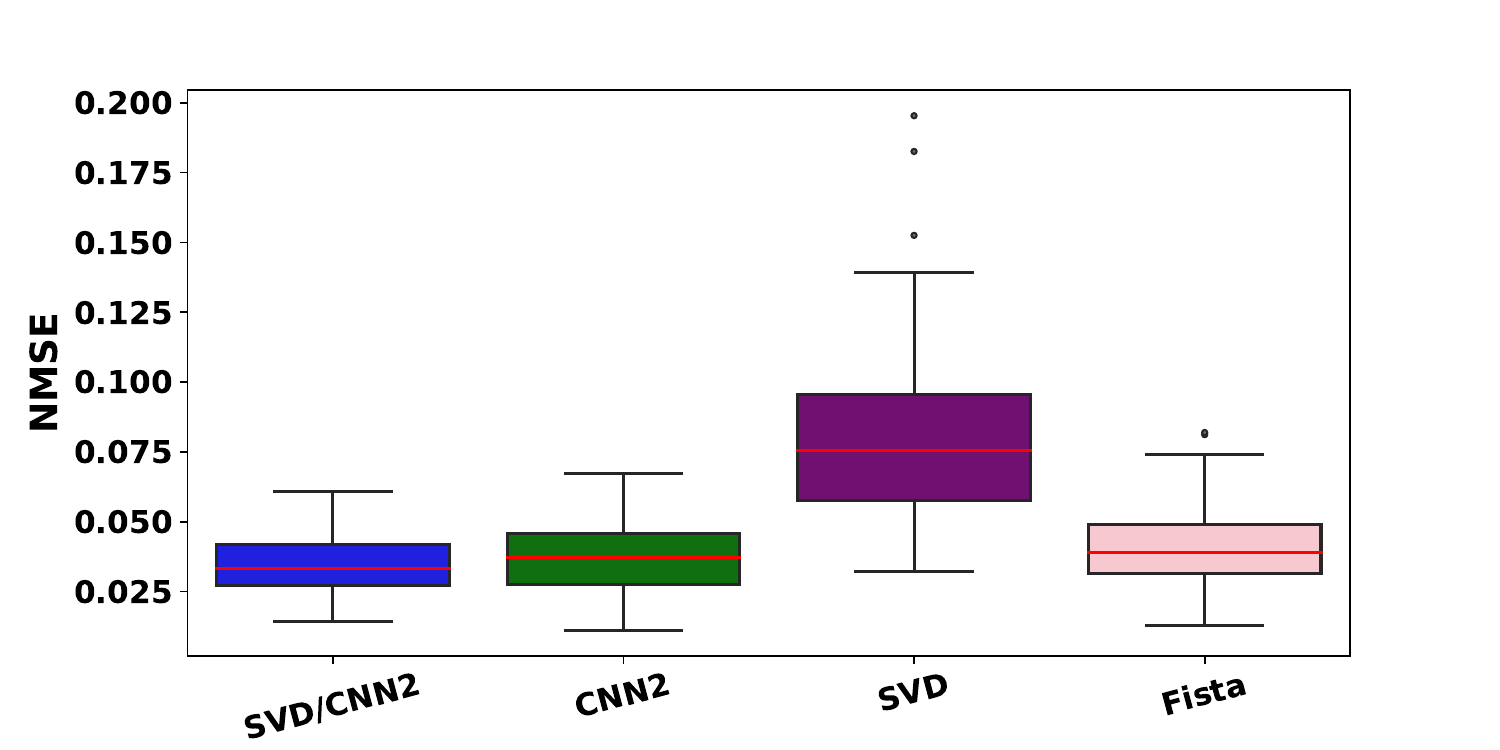}}}
    \qquad
      \hfil
    \subfloat[]{{\includegraphics[width=0.8\linewidth]{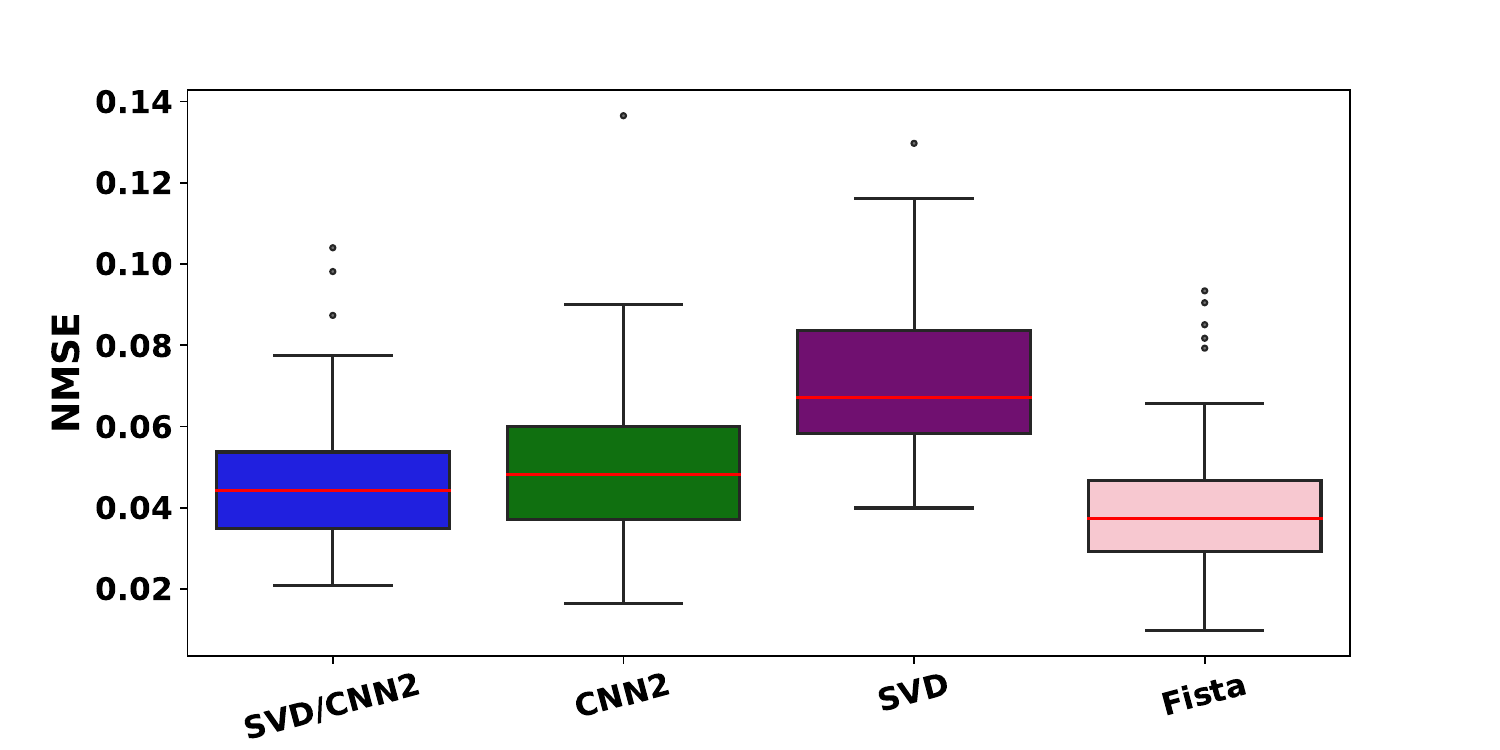}}}
    \qquad
      \hfil
    \subfloat[]{{\includegraphics[width=0.8\linewidth]{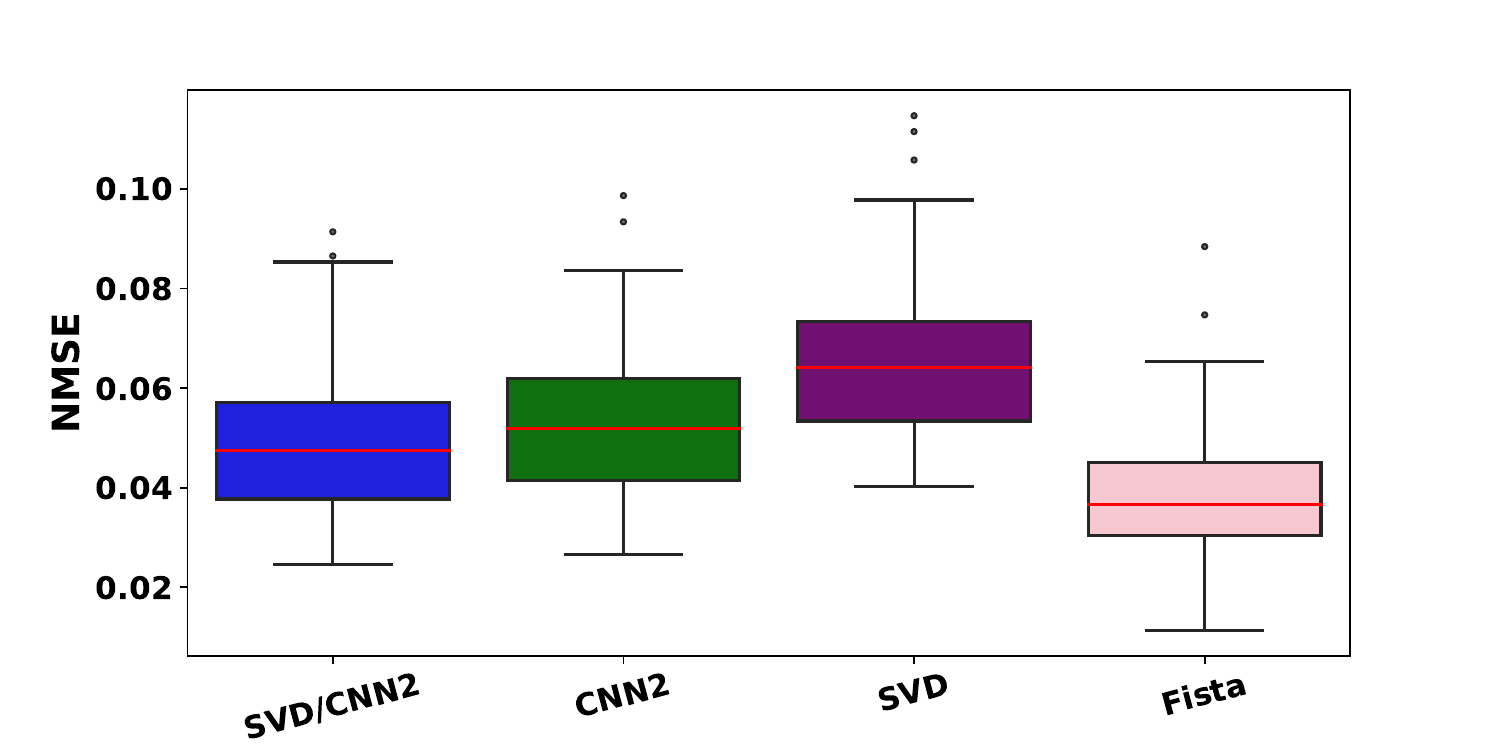} }}
          \hfil
    \subfloat[]{{\includegraphics[width=0.8\linewidth]{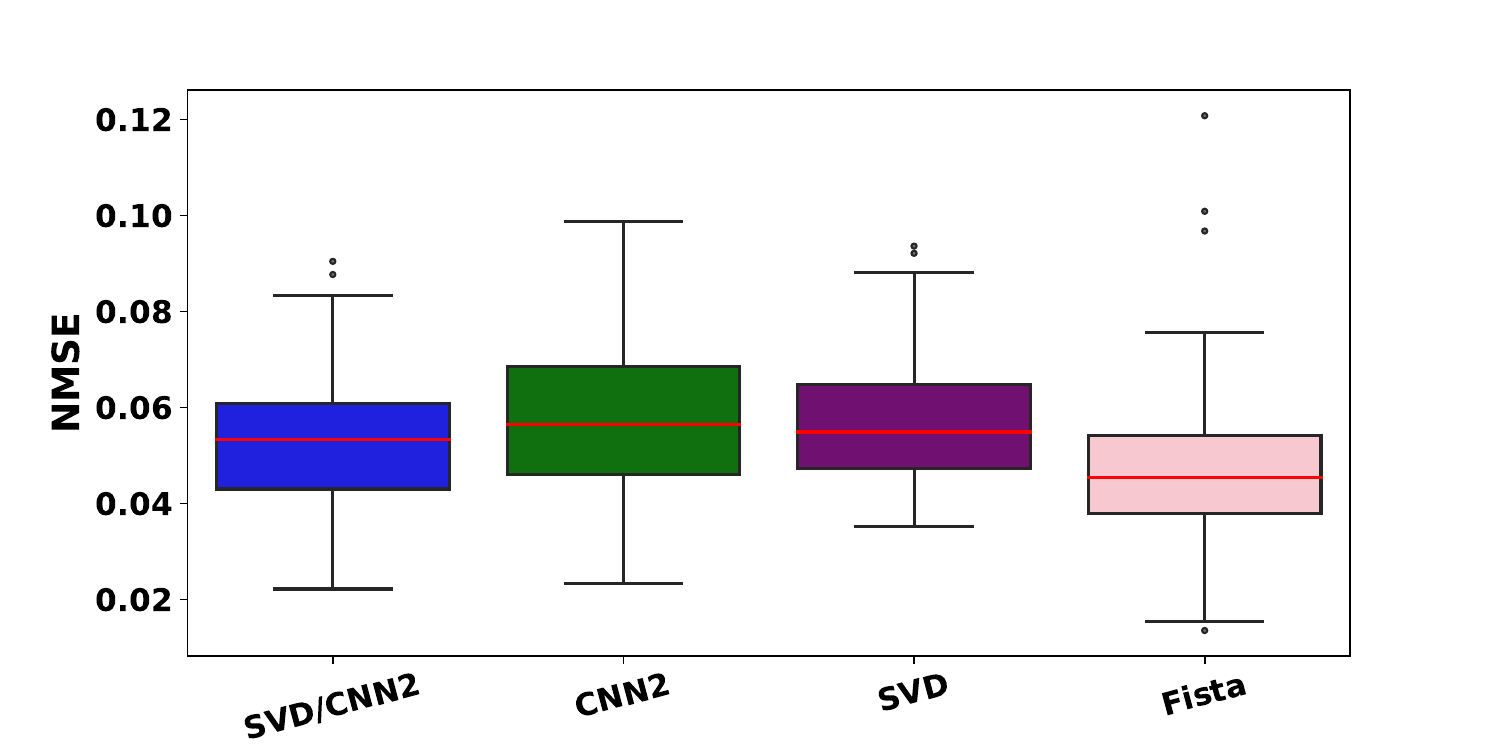} }}
\caption{NMSE of all model for varying K-edges (1-5 K-edge). a) $D_{3E,1K}$, b) $D_{3E,2K}$, c)$D_{3E,3K}$, d) $D_{5E,5K}$}%
\label{fig:all_kedges}%
\end{figure}

To see how the performance of our methods changes when the data has more materials with K-edges in their absorption spectra, we plot the NMSE of the datasets  ($D_{3E,1K}$,  $D_{3E,2K}$, $D_{3E,3K}$, and $D_{5E,5K}$) in Fig. \ref{fig:all_kedges} for the SVD/CNN2, the 5-dimensional CNN2, the 5-dimensional SVD and Fista models. Whilst there is a decrease in the performance of the non-linear models if we increase the number of elements with K-edges, their relative performance is consistent, with Fista, SVD/CNN2  and CNN2 working better than the 5-dimensional SVD model in general.

\begin{figure}
\center 
    \subfloat[]{{\includegraphics[width=0.8\linewidth]{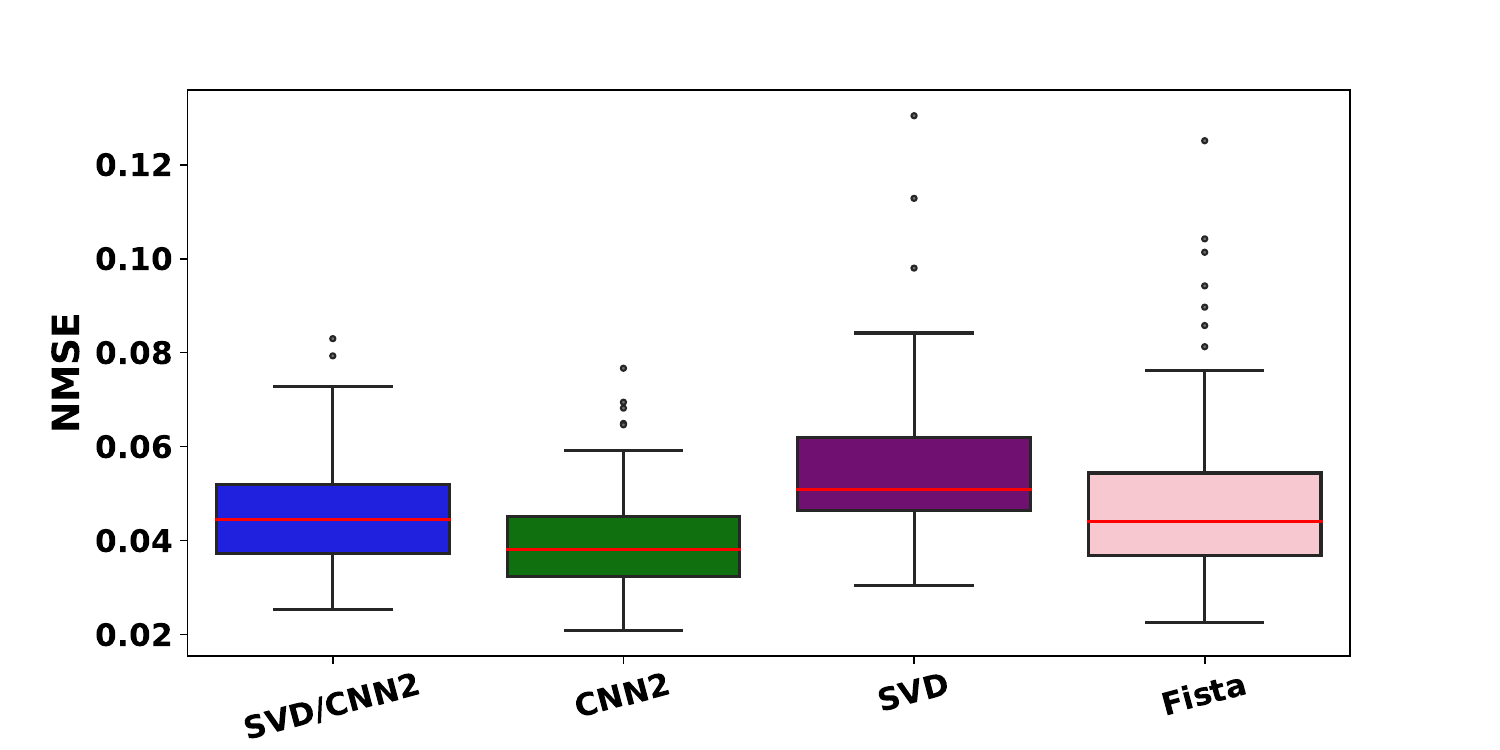}}}
    \qquad
      \hfil
    \subfloat[]{{\includegraphics[width=0.8\linewidth]{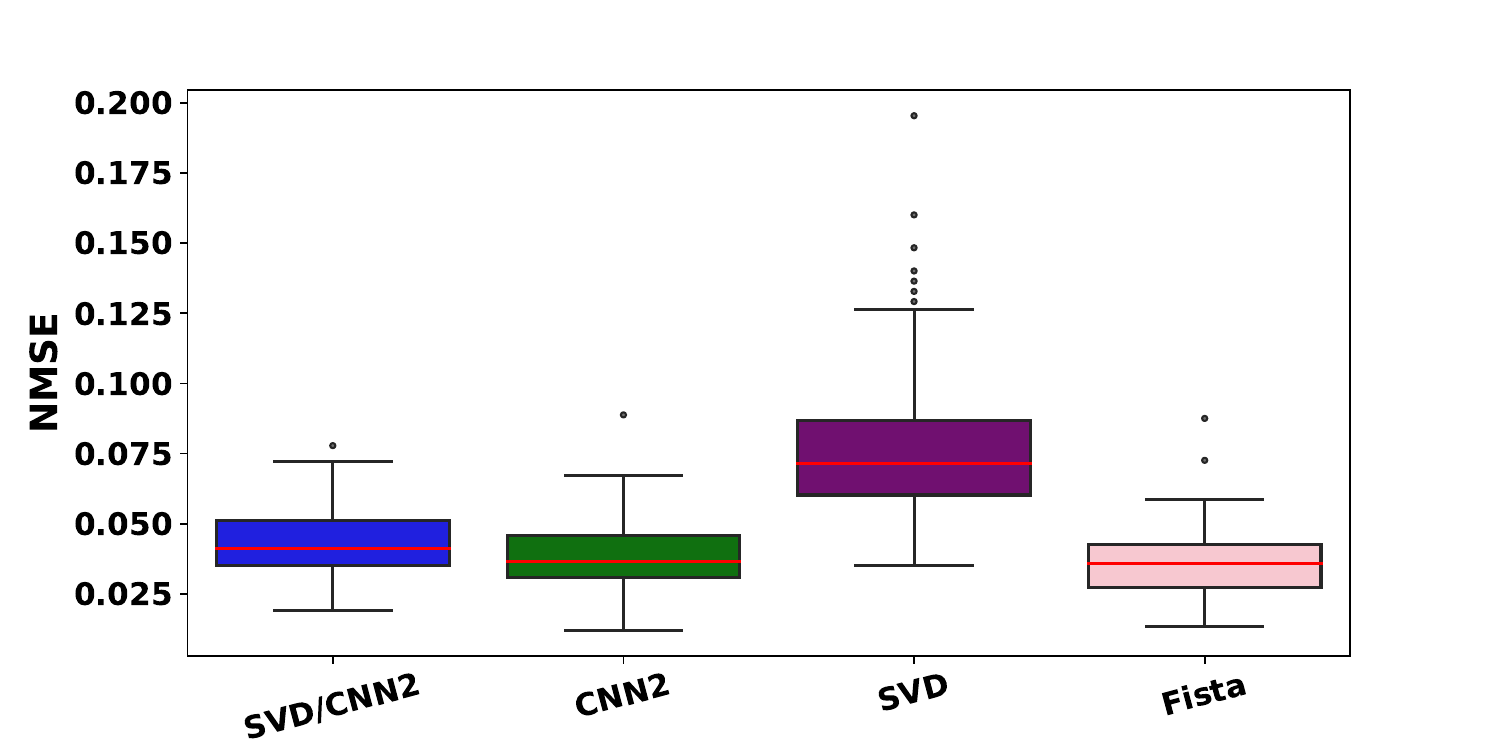}}}

\caption{NMSE of all models trained with five element dataset. a)$D_{5E,5K}$, b)$D_{2E,2K}$ }%
\label{fig:new_training}%
\end{figure}

We also trained our best architectures (SVD/CNN2 and CNN2), 5-dimensional SVD and Fista with two different dataset of 5 materials each to see if their performance depended on the training sets. We here used $D_{5E}$ and $D_{5E,0K}$. The training in this experiment is the same as in the previous one; the only difference is the dataset used for training. After the training, these architectures were tested with $D_{2E,2K}$ and $D_{5E,5K}$. Figures \ref{fig:new_training}a and \ref{fig:new_training}b demonstrate the NMSE result of this experiment, and our models (SVD/CNN2, CNN2 and Fista) still have lower errors than the traditional models.

\section {Discussion and Conclusions}
\label{sec: conclusions and discussion}

Accurate and precise modelling of the X-ray absorption spectrum of objects has been important for reducing image artefacts \cite{blumensath2015non}, estimating material distributions within the object\cite{busi2019enhanced}, and constraining the ill-conditioned inverse problems \cite{kumrular2022multi} that arise in
several spectral imaging methods. In this paper, we considered non-linear models of the energy-dependent X-ray absorption spectrum for all possible materials. We introduced a novel non-linear model, consisting of a linear SVD and a deep learning-based approach, that accurately represents the LACs of K-edge-containing materials using several parameters. Furthermore, we evaluated the performance of different deep learning architectures, traditional linear models, and a sparse model for various simulated objects.

As seen in Fig. \ref{fig:two_element}, all complex architectures (except SVD/FCNN1 and FCNN1), and the Fista model have a lower approximation error than the best linear model (5-dimensional SVD).  Crucially, the traditional linear model has almost the same error, which is 5\% higher than the SVD/FCNN1 and FCNN1 models. This primarily shows that a non-linear model is useful for modelling K-edges. Furthermore, this result suggests that more layers should be used while designing the deep learning architectures for modelling. The last and most important result is that the SVD/CNN2 and CNN2 architectures showed the best performance compared to other architectures in the experiment with the $D_{2E}$ test dataset.

The result of experiments with data with $D_{3E,131}$ dataset showed that our models have the same sensitivity even for finer energy resolution. Interestingly, it shows that if objects with finer resolution have a K-edge in their absorption spectra, the 5-dimensional SVD approach cannot capture it. It can be seen in Fig \ref{fig:131 energy level}, the SVD model has a higher error (10\%) than all other models. For computational efficiency, we did not conduct any further experiments with the 131 energy level dataset, even though our models achieved better performance.

For objects whose K-edge in the X-ray absorption spectrum lies outside of the considered energy range, there is some loss in the SVD/autoencoder approach, as can be seen in Fig.\ref{fig:3element_without_K_edge}. The main reason for this is that we have not trained the autoencoder part in the SVD/autoencoder model with non-K-edge materials. We trained the non-linear step in this model with the residual error (i.e. to model the K-edges), whilst the linear step is trained to model the non-K-edge X-ray absorption spectrum. Since the training methods used to model the non-K-edge X-ray absorption spectrum are not the same (such as non-linear and linear), this is likely to affect the performance of our approach. However, the errors are lower than the best linear model in the SVD/autoencoder, the 5-dimensional autoencoder and the Fista model (error value below 2\% for CNN2, below 4\% for SVD/CNN2 and Fista). Interestingly, the best linear model has a higher error than the other model, even for objects that do not contain K-edges in the X-ray absorption spectrum. Although traditional models are used to model the X-ray absorption spectra of non-K-edge materials in the selected energy range in the literature, these results show that our models can also be used for these spectra.

All experiments with objects with various numbers of K-edges in the X-ray absorption spectrum suggest that our models can be more accurate than the traditional model.
Furthermore, the error in the SVD/autoencoder and the 5-dimensional autoencoder model have increased when the number of K-edges in the X-ray absorption spectrum is increased, as seen in Fig. \ref{fig:all_kedges}a and \ref{fig:all_kedges}b \ref{fig:all_kedges}c \ref{fig:all_kedges}d. Interestingly, the errors of the Fista model for all experiments nearly stayed the same. The reason for this is that there is no training step in the Fista model (apart from fitting the sparsity parameter). Crucially, with the five-element dataset test (as shown in Fig. \ref{fig:new_training}), we found that our models work better than the traditional model, even when trained with more complex datasets.

Our experimental results indicate that using the SVD/autoencoder model approach has significant advantages in the representation of the X-ray absorption spectrum of high atomic number materials compared to the linear model.  In addition, the 5-dimensional autoencoder method has been experimentally shown to work better than traditional linear methods for non-K-edge materials (low atomic number materials) and also complex datasets. Whilst the Fista model did not show good performance for objects that don't have a K-edge, it has good accuracy for objects that have K-edges. The overall utility of our approach lies in that exploring the so-called low-dimensional representation of the X-ray absorption spectrum can be a valuable tool for analyzing the information on the scanned material.

\bibliographystyle{IEEEtran}
\bibliography{cas-refs}

\end{document}